\documentclass[twoside,11pt]{article}
\usepackage{jmlr2e}
\usepackage{xcolor}
\usepackage{amsmath}
\usepackage{booktabs}
\usepackage [english]{babel}
\usepackage [autostyle, english = american]{csquotes}
\MakeOuterQuote{"}
\usepackage{todonotes}
\RequirePackage{algorithm}
\RequirePackage{algorithmic}

\ShortHeadings{Spectral Gap Regularization of Neural Networks}{Tam and Dunson}
\firstpageno{1}

\begin{document}

\title{Spectral Gap Regularization of Neural Networks}

\author{\name Edric Tam \email edric.tam@duke.edu \\
       \addr Department of Statistical Science\\
       Duke University\\
       Durham, NC 27705, USA
       \AND
       \name David B. Dunson \email dunson@duke.edu \\
       \addr Department of Statistical Science\\
       Duke University\\
       Durham, NC 27705, USA}

\editor{N/A}

\maketitle

\begin{abstract}
We introduce Fiedler regularization, a novel approach for regularizing neural networks that utilizes spectral/graphical information. Existing regularization methods often focus on penalizing weights in a global/uniform manner that ignores the connectivity structure of the neural network. We propose to use the Fiedler value of the neural network's underlying graph as a tool for regularization. We provide theoretical motivation for this approach via spectral graph theory. We demonstrate several useful properties of the Fiedler value that make it useful as a regularization tool. We provide an approximate, variational approach for faster computation during training. We provide an alternative formulation of this framework in the form of a structurally weighted $\text{L}_1$ penalty, thus linking our approach to sparsity induction. We provide uniform generalization error bounds for Fiedler regularization via a Rademacher complexity analysis. We performed experiments on datasets that compare Fiedler regularization with classical regularization methods such as dropout and weight decay. Results demonstrate the efficacy of Fiedler regularization. This is a journal extension of the conference paper by \cite{tam2020fiedler}.
\end{abstract}


\section{Introduction}

Neural networks (NNs) are important tools with many applications in various machine learning domains such as computer vision, natural language processing and reinforcement learning. NNs have been very effective in settings where large labeled datasets are available. Empirical and theoretical evidence has pointed to the ever-increasing capacity of recent NN models, both in depth and width, as an important contributor to their modeling flexibility and success. To ameliorate any potential overfitting that comes with these flexible models, a wide range of techniques for regularizing NNs have been developed. These techniques often regularize the network from a global/uniform perspective, e.g. weight decay \citep{krogh1992simple}, $\text{L}_1$/$\text{L}_1$ penalization of weights, dropping nodes/weights in a Bernoulli manner with uniform probability across units/layers \citep{hinton2012improving, srivastava2014dropout,wan2013regularization}, or stopping training early. These commonly used approaches ignore the NN's underlying graphical structure, which can provide valuable connectivity information for regularization.

Existing feedforward NN architectures, e.g. multi-layer perceptrons, frequently employ fully connected layers that lead to many redundant paths between nodes of the network. These redundant connections can contribute to over-fitting through the phenomenon of co-adaptation, where weights become dependent on one another, leading to highly correlated behavior amongst different hidden units \citep{hinton2012improving}. Empirical work has shown that dropping weights and nodes randomly during training can significantly improve test performance by reducing co-adaptation \citep{hinton2012improving, srivastava2014dropout,wan2013regularization}.

One natural alternative to these approaches is to take the graph structure of the NN into consideration during regularization. In this work, we would like to regularize the NN through reducing co-adaptation and penalizing extraneous connections in a way that respects the NN's graphical/connectivity structure. We introduce Fiedler regularization, borrowing from advances in spectral graph theory \citep{godsil2013algebraic, chung1997spectral, Spielman2019Algebraic}. The Fiedler value of a connected graph, denoted $\lambda_2$, also known as the algebraic connectivity, is the second smallest eigenvalue of the graph's Laplacian matrix. The magnitude of $\lambda_2$ characterizes how well connected a graph is: if the Fiedler value is small, then the graph is close to disconnected. By adding the Fiedler value as a penalty term to the loss function during training, we can penalize the connectedness of the NN and reduce co-adaptation while taking into account the graph's connectivity structure.

We also exploit several useful characteristics of the Fiedler value. We show that the Fiedler value is a concave function on the sizes of the NN's weights. This allows us to draw connections to the literature on folded-concave penalties in variable selection, which are widely adopted in statistics to reduce bias in the penalized regression setting. We additionally show that the Fiedler value's gradient with respect to the network's weights admits a closed form expression, which gives insight into the behavior of Fiedler regularization.

In practice, for larger networks, to speed up computation, we propose a variational approach, which replaces the original Fiedler value penalty term by a quadratic form of the graph Laplacian. When used together with the so called Laplacian test vectors, such a Laplacian quadratic form provides a sharp upper bound of the Fiedler value. This variational approximation allows for substantial speedups during training. We give an alternative but equivalent formulation of the variational penalty in terms of a structurally weighted $\text{L}_1$ penalization, where the weights depend on the (approximate) second eigenvector of the graph Laplacian. This $\text{L}_1$ formulation allows us to link Fiedler regularization to sparsity induction, similar to the parallel literature in statistics \citep{tibshirani1996regression, zou2006adaptive}. To provide theoretical guarantees of such an approach, we characterize how Fiedler regularization reduces the Rademacher complexity of neural networks. This leads to uniform risk upper bounds for our approach, which sheds light on generalization performance.

There has been prior work on using the Laplacian structure of the input data to regularize NNs \citep{kipf2016semi, jiang2018graph, zeng2019deep}. There has also been recent work on understanding regularization on NNs in Bayesian settings \citep{vladimirova2018understanding, polson2018posterior}. These approaches do not consider the graphical connectivity structure of the underlying NN. One of the main conceptual contribution of this work is to use the graphical/connectivity structure of a model to regularize itself. After the publication of the conference version of this article, several related directions of research have been pursued.  \citep{carmichael2021folded} adopted a similar folded concave Laplacian spectral (FCLS) penalty in the block sparsity estimation setting in statistics. \citep{arbel2021bayesian} adopted a related Fiedler prior in the Bayesian block-diagonal graphical models setting. In related Laplacian regularization problems considered in \citep{tuck2019distributed} and \citep{carmichael2021folded}, majorization-minization procedures are proposed to optimize the Laplacian regularization term. However, in the context of neural networks, majorization-minimization is not a computationally feasible method for optimization, in part due to the deep and high-dimensional nature of the model as well as the highly non-convex nature of the loss function.

The main contributions of this work include: (1) to the authors' best knowledge, this is the first application of spectral graph theory and Fiedler values in regularization of NNs via their own underlying graphical/connectivity structures. (2) We give practical and fast approaches for Fiedler regularization, along with strong theoretical guarantees and experimental performances. This is a journal extension of the conference paper \citep{tam2020fiedler}. The journal version extends the conference paper by providing new generalization error guarantees via a Rademacher complexity analysis as well as a new connection to the variable selection literature via the folded-concave penalty interpretation.

\section{Spectral Graph Theory}
\subsection{Setup and Background}
Let $W$ denote the set of weights of a feedforward NN $\mathbf{f}$. We denote $\mathbf{f}$'s underlying graph structure as $G$. We would like to use structural information from $G$ to regularize $\mathbf{f}$ during training. Feedforward NNs do not allow for self-loops or recurrent connections, hence it suffices that $G$ be a finite, connected, simple, weighted and undirected graph in this setting. Such a graph $G$ can be fully specified by a triplet $(V, E, |W|)$. Here the vertex set $V$ of $G$ corresponds to all the units (including units in the input and output layers) in the NN $\mathbf{f}$, while the edge set $E$ of $G$ corresponds to all the edges in $\mathbf{f}$. For our purposes of regularization, $G$ is restricted to have non-negative weights $|W|$, which are taken to be the absolute value of the corresponding weights $W$ in the NN $\mathbf{f}$. Throughout the paper we use $n$ to denote the number of vertices in $G$.  $|W|$ and $E$ can be jointly represented by an $n \times n$ weighted adjacency matrix $\mathbf{|W|}$, where $\mathbf{|W|}_{ij}$ is the weight on the edge $(i,j)$ if vertices $i$ and $j$ are connected, and 0 otherwise. The degree matrix $\mathbf{D}$ of the graph is a $n \times n$ diagonal matrix, where $\mathbf{D}_{ii} = \sum_{j = 1}^n \mathbf{|W|}_{ij}$. The Laplacian matrix $\mathbf{L}$, which is a central object of study in spectral graph theory, is defined as the difference between the degree and the adjacency matrix, i.e. $\mathbf{L} = \mathbf{D}-\mathbf{|W|}$. In certain contexts where we would like to emphasize the dependency of $\mathbf{L}$ on the particular graph $G$ or the weights $|\mathbf{W}|$, we adopt the notation $\mathbf{L}_G$ or $\mathbf{L}_{|\mathbf{W}|}$. We use $[M]$ to denote the set $\{1,2,\cdots, M\}$ for positive integer $M$. Throughout the paper, eigenvalues are real-valued since the matrices under consideration are symmetric. We adopt the convention where all eigenvectors are taken to be unit vectors. We order the eigenvalues in ascending order, so $\lambda_i \leq \lambda_j$ for $i < j$. When we want to emphasize $\lambda_i$ as a function of the weights, we use $\lambda_i(\mathbf{|W|})$. We use $\mathbf{v}_i$ to denote the corresponding eigenvector for $\lambda_i$. We use $n(S)$ to denote the cardinality of a given set $S$.

\subsection{Graph Laplacian}
The Laplacian matrix encodes much information about the structure of a graph. Laplacian eigenvalues and eigenvectors are widely used in tasks such as spectral clustering \cite{von2007tutorial, ng2001spectral} and manifold learning \cite{belkin2003laplacian}, precisely because they capture valuable graph connectivity information. One particularly useful characterization of the graph Laplacian that we use heavily is the so called Laplacian quadratic form \citep{batson2012twice, Spielman2019Algebraic, chung1997spectral}, defined below.
\\\\{\bf Definition 2.2.1 (Laplacian quadratic form)} {\it Given a graph $G = (V, E, |W|)$ with Laplacian matrix $\mathbf{L}_G$, define its Laplacian quadratic form $Q_G: \mathbb{R}^{|V|} \to \mathbb{R}^+$ as
\[
	Q_G(\textbf{z}) := \textbf{z}^T \mathbf{L}_G \textbf{z} = \sum_{(i,j) \in E} \mathbf{|W|}_{ij} (\textbf{z}(i) - \textbf{z}(j))^2,
\]
where $\textbf{z}(k)$ denotes the k\textsuperscript{th} entry of the vector $\textbf{z} \in \mathbb{R}^{|V|}$.
} \\\\
The Laplacian quadratic form demonstrates how a graph's boundary information can be recovered from its Laplacian matrix. One can think of $\textbf{z}$ as a mapping that assigns to each vertex a value. If we denote the characteristic vector of a subset of vertices $S \subset V$ as $\textbf{1}_S$, i.e. $\textbf{1}_S (i) = 1$ if $i \in S$ and $\textbf{1}_S (i) = 0$ otherwise, and apply it to the Laplacian quadratic form, we obtain $\textbf{1}_S^T \mathbf{L}_G \textbf{1}_S = \sum_{(i,j) \in E, i\in S, j \not \in S} \mathbf{|W|}_{ij}$. This expression characterizes the size of the graph cut $(S,  V-S)$, which is the sum of the weights of edges crossing the boundary between $S$ and $V-S$. The size of any graph cut can therefore be obtained by application of the corresponding characteristic vector on the Laplacian quadratic form.

\subsection{Edge Expansion and Cheeger's Inequality}
The above discussion on the sizes of graph cuts is highly related to our study of regularizing a NN. Reducing the sizes of graph cuts in a NN would imply reducing the NN's connectivity and potential co-adaptation. A related but more convenient construct that captures this notion of boundary sizes in a graph is the graph's edge expansion (also known as the Cheeger constant or the isoperimetric constant), which can be informally thought of as the smallest "surface-area-to-volume ratio" achieved by a subset of vertices not exceeding half of the graph.
\\\\{\bf Definition 2.3.1 (Edge expansion of a graph)} {\it The edge expansion $\phi_G$ of a graph $G = (V, E, |W|)$ is defined as
\[
\phi_G = \min_{S \subset V, n(S) \leq \frac{n(V)}{2}} \frac{\sum_{i \in S, j \not \in S} \mathbf{|W|}_{ij}}{n(S)},
\]
where $n(S)$ denotes the number of vertices in $S$.
}
\\\\
Observe that the term in the numerator characterizes the size of the graph cut $(S, V-S)$, while the denominator normalizes the expression by the number of vertices in $S$. The edge expansion is then taken to be the smallest such ratio achieved by a set of vertices $S$ that has cardinality at most half that of $V$. One can think of the edge expansion of a graph as characterizing the connectivity bottleneck of a graph. It is highly related to how sparse the graph is and whether there exist nice planar embeddings of the graph \citep{hall1970r}.

\begin{figure}
    \centering
    \includegraphics[width = 0.8\textwidth]{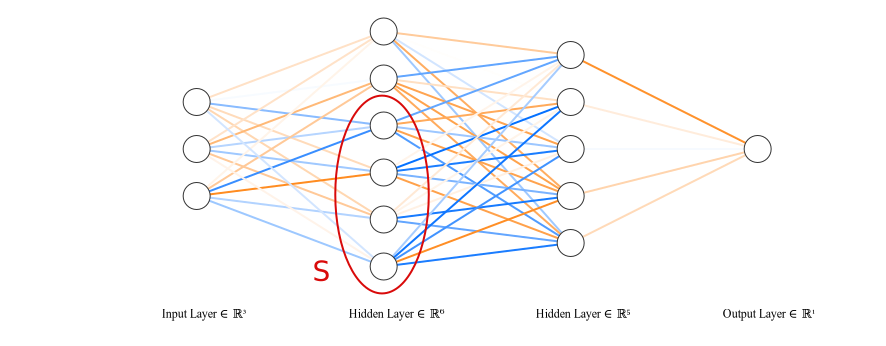}
    \caption{Graphical illustration of graph cut $(S, S^C)$ on neural networks. A subset of vertices $S$ are circled in red, and the size of the graph cut $(S, S^C)$ is the sum of the sizes of the weights of all the edges that intersect the red boundary. Weights are colored by their signs (orange for positive, blue for negative) and their opacity determined by the absolute values. }
    \label{fig:my_label}
\end{figure}

We would like to control the edge expansion of the NN's underlying graph for regularization. It is well known that direct computation of the edge expansion is a hard problem \cite{garey1974some} due to the combinatorial structure. Instead, we control the edge expansion indirectly through the Fiedler value $\lambda_2$, the second smallest eigenvalue of the graph's Laplacian $\mathbf{L}_G$.  $\lambda_2$ is related to the edge expansion of the graph through Cheeger's inequality \citep{Spielman2019Algebraic, chung1997spectral, godsil2013algebraic}.
\\\\ {\bf Proposition 2.3.2 (Cheeger's inequality) } {\it Given a graph $G = (V, E, |W|)$, the edge expansion $\phi_G$ is upper and lower bounded as follows:
\[
	\sqrt{2d_{\max}(G)\lambda_2}\geq \phi_G \geq \frac{\lambda_2}{2},
\]
where $d_{\max}(G)$ is the maximum (weighted) degree of vertices in $G$ and $\lambda_2$, the Fiedler value, is the second smallest eigenvalue of $G$'s Laplacian matrix.
} \\\\
Cheeger's inequality originally arose from the study of Riemannian manifolds and was later extended to graphs. There are many versions of Cheeger's inequality depending on the types of graphs and normalizations used. The proofs can be found in many references, including \cite{Spielman2019Algebraic, chung1997spectral, godsil2013algebraic}. These types of Cheeger's inequalities are generally tight asymptotically up to constant factors.

The Fiedler value is also known as the algebraic connectivity because it encodes considerable information about the connectedness of a graph. Cheeger's inequality allows sharp control over the edge expansion of $G$ via the Fiedler value. By making the Fiedler value small, we can force the edge expansion of the graph to be small, thus reducing the connectedness of the graph and potentially alleviating co-adaptation in the NN setting. On the other hand, a large Fiedler value necessarily implies a large edge expansion. This implies that penalization of the Fiedler value during NN training is a promising regularization strategy to reduce connectivity and thus co-adaptation.

\section{Fiedler Regularization}

In this section we introduce Fiedler Regularization of neural networks. We characterize this in a supervised classification setting to illustrate the main ideas, even though the general framework is easily extendable to other settings such as regression and autoencoding. The relevant setup and background is introduced below.

\subsection{Supervised Learning: Classification Setup}
We are given training data $\{(\mathbf{x}_i, y_i)\}_{i = 1}^N$ assumed to be drawn independently from some joint distribution $\mathcal{P}$. Here $i$ indexes the $N$ data points, $\mathbf{x} \in \mathbb{R}^d$ denotes the $d$-dimensional feature vectors, and $y$ denotes the outcomes, which for simplicity of presentation we assume to take values in a binary space $\{0, 1\}$ (i.e. binary classification). Many of our results can be easily generalized to settings such as multi-class classification and regression with straightforward modifications. We aim to find a mapping $\mathbf{f}$ that predicts $y$ through $\mathbf{f}(\mathbf{x})$ so that the expectation of some pre-specified loss $\mathbb{E}_{\mathcal{P}}\mathcal{L}(f(\mathbf{x}), y)$ is minimized. Typical choices of the loss function $\mathcal{L}(\cdot,\cdot)$ for classification include the cross-entropy loss, hinge loss etc, whereas for regression problems the mean squared error is common. The empirical approximation $\mathbb{E}_{\mathcal{P}_N}\mathcal{L}(f(\mathbf{x}), y):= \frac{1}{N} \sum_{i = 1}^N \mathcal{L}(f(\mathbf{x_i}), y_i)$ is used as a proxy for the expected loss during training, where $\mathcal{P}_N$ denotes the empirical measure. It is assumed that new observations in the testing set are sampled independently from the same distribution $\mathcal{P}$ as the training data. We denote the estimator of $\mathbf{f}$ as $ \mathbf{\hat f}$. When we want to emphasize the estimator's dependence on the weights, we use $ \mathbf{ \hat f_W}$.

\subsection{Neural Network Setup}
For Fiedler Regularization, we consider for now only feedforward NNs (see discussion for potential extensions). For feedforward NNs, $\mathbf{\hat f}$ is characterized as a composition of non-linear functions $\{ \mathbf{g}^{(l)}\}_{l = 1}^\Lambda$, i.e. $\mathbf{\hat f}= \mathbf{g}^{(\Lambda)} \circ \mathbf{g}^{(\Lambda-1)} \circ \cdots \circ \mathbf{g}^{(1)}(\mathbf{x})$, where $\Lambda$ is the number of layers in the network. The outputs of the $l$\textsuperscript{th} layer have the form $\mathbf{h}^{(l)} := \mathbf{g}^{(l)}(\mathbf{h}^{(l-1)}) := \sigma^{(l)}(\mathbf{W}^{(l)} \mathbf{h}^{(l-1)} + \mathbf{b}^{(l)})$, where $\sigma^{(l)}$, $\mathbf{W}^{(l)}$ and $\mathbf{b}^{(l)}$ are the activation function, weight matrix and bias of the $l$\textsuperscript{th} layer of the NN, respectively. Note that we use $\mathbf{W}$ to denote the weighted adjacency matrix of the entire neural network encoded as a graph, and $\mathbf{W}^{(l)}$ to denote the matrix encoding the linear transformation applied at the $l$\textsuperscript{th} layer. In essence, each hidden layer first performs an affine transformation on the previous layer's outputs, followed by an element-wise activation that is generally nonlinear. For additional details, see the excellent review by \cite{fan2019selective}.

\subsection{Penalizing with Fiedler Value}
Given the motivations from section 2, we would like to penalize the graph connectivity of the NN during training. In the Fiedler regularization approach, we add $\lambda_2$ as a penalty term to the objective.
\\\\{\bf Definition 3.2.1 (Fiedler Regularization)} {\it During training of the neural network, we optimize the following objective:
\[
	\min_{\mathbf{W}} \mathbb{E}_{\mathcal{P}_N}\mathcal{L}_{\mathbf{W}}+ \delta \lambda_2(\mathbf{|W|}),
\]
where $\lambda_2(\mathbf{|W|})$ is the Fiedler value of the NN's underlying graph, $\mathbf{W}$ is the weight matrix of the NN, $\delta$ is a tuning parameter, and $Y$ and $X$ denote the training labels and training features, respectively.
}
\\\\
We remark that the actual NN $\mathbf{\hat f_W}$ has weights that can be negative, but the regularization term $\lambda_2(\mathbf{|W|})$ only depends on the sizes of such weights, which can be thought of as edge capacities of the underlying graph $G$.

Note that one can generalize the penalty by applying a differentiable function $Q$ to $\lambda_2$. For our discussion below, we focus on the case where no $Q$ is applied in order to focus the analysis on the Fiedler value. However, incorporating $Q$ is straightforward, and desirable properties such as having a closed-form gradient can generally be retained. We also note that, without loss of generality, one can consider the biases of the units in the NN as additional weights with constant inputs, so it is straightforward to include consideration of both biases and weights in Fiedler regularization.

For our purposes, we consider the main parameters of interest to be the weights of the NN. The choice of the activation function(s), architecture, and hyperparameters such as the learning rate or the tuning factor are all considered to be pre-specified in our study. There is a separate and rich literature devoted to methods for selecting activation functions/architectures/hyperparameters that we do not consider here.

It is worth noting that the Fiedler Regularization defined in this section is meant to illustrate the core ideas and motivations in a theoretical and idealized way. In practice, it is computationally intensive to penalize the Fiedler value directly, and for implementation, we refer to the approximate variational approach described in the subsequent section.

\subsection{Properties of the Fiedler Value}

It is instructive to examine some properties of the Fiedler value in order to understand why it is an appropriate tool for regularization.

First, one concern is whether using the Fiedler value as a penalty in the objective would complicate the optimization process. The Fiedler value can be viewed as a root of the Laplacian matrix's characteristic polynomial, which in higher dimensions has no closed-form solution and can depend on the network's weights in a convoluted manner.

To address this concern, the following proposition shows that the Fiedler penalty is a concave function of the sizes of the NN's weights. This shows that when we add the Fiedler penalty to deep learning objectives, which are typically highly non-convex, we are not adding substantially to the optimization problem's difficulty.
\\\\{\bf Proposition 3.3.1 (Concavity of Fiedler Value)} {\it The function $\lambda_2(|W|)$
is a concave function of the sizes of the NN's weights $|W|$. \\
}
\noindent{\it Proof:}{ Since the Fiedler value $\lambda_2$ is just the second smallest eigenvalue of the Laplacian, and we know that the first eigenvector of the Laplacian must be constant, we can consider $\lambda_2$'s Rayleigh-Ritz variational characterization as follows:
$$\lambda_2(\mathbf{|W|}) = \inf_{||\textbf{u}|| = 1, \textbf{u}^T\textbf{1} = 0} \textbf{u}^T\mathbf{L}\textbf{u}$$
$$= \inf_{||\textbf{u}|| = 1, \textbf{u}^T\textbf{1} = 0} \sum_{(i,j) \in E} \mathbf{|W|}_{ij}(\textbf{u}(i) - \textbf{u}(j))^2$$
Note that this is a pointwise infimum of a linear function of $\mathbf{|W|}_{ij}$. Since linear functions are concave (and convex), and the pointwise infimum preserves concavity, we have that $\lambda_2$ is a concave function of the sizes of the weights. \BlackBox
}
\\\\This is related to Laplacian eigenvalue optimization problems and we refer to \cite{boyd2006convex} and \cite{sun2006fastest} for a more general treatment.

An immediate corollary of Proposition 3.3.1 is that the Fiedler value is a folded-concave penalty with respect to the weights $W$ of the neural network. Typical sparsity-inducing methods with $\text{L}_1$ penalties, such as the least absolute shrinkage and selection operator (LASSO) in statistics \citep{tibshirani1996regression}, can suffer from large biases. Methods with folded-concave penalties, such as the Smoothly Clipped Absolute Deviation (SCAD) \citep{fan2001variable} penalty and the minimax concave penalty (MCP) \citep{zhang2010nearly}, have been shown to reduce the biases of $\text{L}_1$-based sparsity-inducing methods. This shape of folded-concave penalty functions leads to the property that smaller weights are going to be penalized relatively more than the larger weights. In the case of Fiedler regularization, the weights of neural network that have "low connectivity" are penalized more. We can further understand this phenomenon via analyzing the gradient of the Fiedler value with respect to the neural network's weights.

\subsection{Closed-form Expression of Gradient}
In all except the most simple of cases, optimizing the loss function $\text{min}_{\mathbf{W}} \mathcal{L}(Y, \mathbf{f_W}(X))$ is a non-convex problem. There are a variety of scalable, stochastic algorithms for practical optimization on such objectives. Virtually all of the widely used methods, such as stochastic gradient descent (SGD) \citep{ruder2016overview}, Adam \citep{kingma2014adam}, Adagrad \citep{duchi2011adaptive}, RMSProp \citep{graves2013generating} etc, require computation of the gradient of the objective with respect to the parameters.

We provide a closed-form analytical expression of the gradients of a general Laplacian eigenvalue with respect to the entries of the Laplacian matrix. From that, as a straightforward corollary, a closed-form analytical expression of the Fiedler value's gradient is obtained.
\\\\ {\bf Proposition 3.4.1 (Gradient of Laplacian Eigenvalue) } {\it Assuming that the eigenvalues of the Laplacian $\mathbf{L}$ are not repeated, the gradient of the $k$\textsuperscript{th} smallest eigenvalue $\lambda_k$ with respect to $\mathbf{L}$'s $(ij)$\textsuperscript{th} entry $\mathbf{L}_{ij}$ can be analytically expressed as
$$\frac{d\lambda_k}{d\mathbf{L}_{ij}} = \textbf{v}_k(i)\times \textbf{v}_k(j),$$
where $\textbf{v}_k(i)$ denotes the $i$\textsuperscript{th} entry of the $k$\textsuperscript{th} eigenvector of the Laplacian. We adopt the convention in which all eigenvectors under consideration are unit vectors. \\\\
}
\noindent {\it Proof:}
{To compute $\frac{d\lambda_k}{d\mathbf{L}_{ij}}$, note that since the Laplacian matrix is symmetric, all eigenvalues are real. By assumption, the eigenvalues are not repeated. Under this situation, there is an existing closed-form formulae (see \citep{petersen2008matrix}): $\textbf{v}_k^T(\partial \mathbf{L})\textbf{v}_k = d\lambda_k$. Specializing to individual entries, we get $\frac{d\lambda_k}{d\mathbf{L}_{ij}} = \textbf{v}_k(i)\times \textbf{v}_k(j)$. \BlackBox
}
\\\\From the above, we can easily obtain an analytical expression for the gradient of the Fiedler value with respect to the weights of the neural network.
\\\\{\bf Corollary 3.4.2 (Gradient of Fiedler value with respect to weights) } {\it Under the same assumptions of Proposition 3.4.1, as an immediate special case, the gradient of the Fiedler value $\lambda_2$ can be expressed as:
$$\frac{d\lambda_2}{d\mathbf{L}_{ij}} = \textbf{v}_2(i) \times \textbf{v}_2(j)$$
Since $\mathbf{L}_{ij} = -\mathbf{|W|}_{ij}$ for $i \not = j$, this yields the following:
$$\frac{d\lambda_2}{d\mathbf{\mathbf{|W|}}_{ij}} = -\textbf{v}_2(i) \times \textbf{v}_2(j)$$
}
\\Under the additional assumption that the weights under consideration are non-zero, one can remove the absolute value mapping in the gradient by a simple application of the chain rule.

Our assumption that the eigenvalues are not repeated generally holds in the context of NNs. The weights in a NN are usually initialized as independent draws from certain continuous distributions, such as the uniform or the Gaussian. Repeated Laplacian eigenvalues often occur when there are strong symmetries in the graph. Such symmetries are typically broken in the context of NNs since the probability of different weights taking the same non-zero value at initialization or during training is negligible.

The reason we develop the above analytical form for the gradient of the Fiedler value is not for direct optimization of real life neural networks: recomputing the second Laplacian eigenvector at every iteration of optimization (e.g. in SGD) would be prohibitively expensive for all but the smallest neural networks. Rather, we use the analytical form of the gradient to shed light on what Fiedler Regularization achieves theoretically. For actual computation, we propose a tight variational approximation in the next section which works well in practice.

There is a large literature on spectral clustering that analyzes the interpretation of the entries of the second Laplacian eigenvector. In particular, if vertices $i$ and $j$ are poorly connected to each other (i.e. they belong to different "clusters"), then $\textbf{v}_2(i) \times \textbf{v}_2(j)$ would be a very small/negative real number. This indicates that the derivative $\frac{d\lambda_2}{d\mathbf{\mathbf{|W|}}_{ij}} = -\textbf{v}_2(i) \times \textbf{v}_2(j)$ is larger for vertices $i$ and $j$ that are less well connected, which in turn implies that edges/weights between poorly connected vertices gets penalized most under Fiedler regularization.

\section{Variational, Approximate Approach for Computational Speedup}
We have given theoretical motivation and outlined the use of the Fiedler value as a tool for NN regularization. However, typical matrix computations for eigenvalues and eigenvectors are of order $O(n^3)$, where $n = |V|$. This can be computationally prohibitive for moderate to large networks. Even though there exist theoretically much more efficient algorithms to approximate eigenvalues/eigenvectors of graph-related matrices, in practice computing the Fiedler value in every iteration of training can prove costly. To circumvent this issue, we propose an approximate, variational approach to speed up the computation to $O(|E|)$, where $|E|$ is the number of edges in the graph. This proposed approach can be readily implemented in popular deep learning packages such as Tensorflow and PyTorch.

We make the following observations. First, for the purpose of regularizing a NN, we do not need the exact Fiedler value of the Laplacian matrix. A good approximation suffices. Second, there is no need to update our approximate $\lambda_2$ at every iteration during training. We can set a schedule to update our $\lambda_2$ approximation periodically, say once every $100$ iterations. Both of these observations allow for substantial speedups in practice. An outline of the pseudo-code for this approximate, variational approach is provided in Algorithm 1.

To obtain an approximate Fiedler value, we use a special type of Laplacian quadratic form involving the so called test vectors.
We additionally provide a perturbation bound that gives justification for periodically updating the Fiedler value.

\subsection{Rayleigh Quotient Characterization of Eigenvalues}
We can closely upper-bound the Fiedler value via the notion of test vectors \citep{Spielman2019Algebraic}, which depends crucially upon the Rayleigh quotient characterization of eigenvalues.
\\\\{\bf Proposition 4.1.1 (Test Vector Bound) } {\it For any unit vector $\textbf{u}$ that is perpendicular to the constant vector $\textbf{1}$, we have: $$\lambda_2 \leq \textbf{u}^T\mathbf{L}\textbf{u}$$  Any such unit vector is called a test vector. Equality is achieved when $\textbf{u} = \textbf{v}_2$. \\\\
}
\noindent {\it Proof:}
{The Laplacian matrix of a non-negatively weighted graph is symmetric and positive semidefinite. Thus all eigenvalues are real and non-negative. In particular, the smallest eigenvalue of the Laplacian is 0, with the constant vector being the first eigenvector. For a connected graph, the second smallest eigenvalue, which is the Fiedler value, can thus be variationally characterized via the Courant-Fischer theorem by $\lambda_2 = \min_{||\textbf{u}|| = 1, \textbf{u}^T \textbf{1} = 0}\textbf{u}^T\mathbf{L}\textbf{u}$. This gives us the desired upper bound. \BlackBox \\\\ }
The above description implies that by appropriately choosing test vectors, we can effectively upper bound the Fiedler value. This in turn implies that during training, instead of penalizing by the exact Fiedler value, we can penalize by the quadratic form upper bound instead.
In other words, we would perform the following optimization,
$$\min_{\mathbf{W}} \mathcal{L}(Y, \mathbf{\hat f_W}(X)) + \delta \mathbf{u}^T\mathbf{L}\mathbf{u}$$
For a given $\mathbf{u}$, this speeds up computation of the penalty term considerably to $O(|E|)$ since $\textbf{u}^T\mathbf{L}\textbf{u} = \sum_{(i,j)\in E}\mathbf{|W|}_{ij}(\textbf{u}(i) - \textbf{u}(j))^2$. The core question now becomes how to choose appropriate test vectors $\mathbf{u}$ that are close to $\textbf{v}_2$.

We propose to initialize training with the exact $\textbf{v}_2$ and recompute/update $\textbf{v}_2$ only periodically during training. For an iteration in between two exact $\textbf{v}_2$ updates, the $\textbf{v}_2$ from the previous update carries over and serves as the test vector for the current iteration. By proposition 4.1.1, this always upper-bounds the true $\lambda_2$.

During training, weights of the NN are updated at each iteration. This is equivalent to adding a symmetric matrix $\mathbf{H}$ to the Laplacian matrix $\mathbf{L}$, which is also symmetric, at every iteration. We can bound the effect of such a perturbation on the eigenvalues via Weyl's inequality \citep{horn2012matrix, horn1998eigenvalue}.
\\\\{\bf Proposition 4.1.2 (Weyl's Inequality)} {\it
Given a symmetric matrix $\mathbf{L}$ and a symmetric perturbation matrix $\mathbf{H}$, both with dimension $n \times n$, for any $1\leq i \leq n$, we have:
$$|\lambda_i(\mathbf{L}+\mathbf{H}) - \lambda_i(\mathbf{L})| \leq ||\mathbf{H}||_{op}$$
where $||\cdot||_{op}$ denotes the operator norm. }
\\\\ Weyl's inequality is a classic result in matrix analysis, and its proof can be found in the excellent reference \citep{horn2012matrix} as a straightforward result of linearity and the Courant-Fischer theorem. As an immediate special case, $|\lambda_2(\mathbf{L}+\mathbf{H}) - \lambda_2(\mathbf{L})| \leq ||\mathbf{H}||_{op}$. Proposition 4.1.2 tells us that as long as the perturbation $\mathbf{H}$ is small, the change in Fiedler value caused by the perturbation is also small. In fact, this shows that the map $\mathbf{L}\to \lambda_2(\mathbf{L})$ is Lipschitz continuous on the space of symmetric matrices. In the context of training NNs, $\mathbf{H}$ represents the updates to the weights of the network during training. This justifies updating $\lambda_2$ only periodically for the purposes of regularization. It suggests that with a smaller learning rate, $\mathbf{H}$ would be smaller, and therefore the change to $\lambda_2$ would also be smaller, and updates of the test vectors can be more spaced apart. On the other hand, for larger learning rates we recommend updating test vectors more frequently.

\begin{algorithm}[tb]
   \caption{Variational Fiedler Regularization with SGD}
   \label{alg:fiedler}
\begin{algorithmic}
   \STATE \noindent {\bf Input:} Training data $\{\mathbf{x}_i, y_i\}_{i = 1}^N$
   \STATE \noindent {\bf Hyperparameters:} Learning rate $\eta$, batch size $m$, penalty parameter $\delta$, updating period $T$
   \STATE \noindent {\bf Algorithm:}
   \STATE Initialize parameters $\mathbf{W}$ of the NN
   \STATE Compute the Laplacian $\mathbf{L_{|W|}}$ of the NN
   \STATE Compute the Fiedler vector $\textbf{v}_2$ of the Laplacian $\mathbf{L_{|W|}}$
   \STATE Set $\textbf{u} \leftarrow \textbf{v}_2$
   \STATE Initialize counter $c = 0$
   \WHILE{Stopping criterion not met}
   \STATE Sample minibatch $\{\mathbf{x}^{(i)}, y^{(i)}\}_{i = 1}^m$ from training set
   \STATE Set gradient $\pmb{\gamma} = \textbf{0}$
   \FOR{ $i=1$ to $m$}
   \STATE Compute gradient
   $\pmb{\gamma} \leftarrow \pmb{\gamma} + \nabla_{\mathbf{W}} \mathcal{L}(\mathbf{\hat{f}_W}(\mathbf{x}^{(i)}), y^{(i)}) + \delta \nabla_{\mathbf{W}}\textbf{u}^T \mathbf{L_{|W|}} \textbf{u}$
   \ENDFOR
   \STATE Apply gradient update  $\mathbf{W} \leftarrow \mathbf{W} - \eta \pmb{\gamma}$
   \STATE Update Laplacian matrix $\mathbf{L_{|W|}}$
   \STATE Update counter $c \leftarrow c + 1$
   \IF{$c$ \text{mod} $T = 0$}
   \STATE Recompute $\textbf{v}_2$ from $\mathbf{L_{|W|}}$
   \STATE Set $\textbf{u} \leftarrow \textbf{v}_2$
   \ENDIF
   \ENDWHILE

\end{algorithmic}
\end{algorithm}

\section{Weighted-$\text{L}_1$ Formulation and Sparsity}
We now expand on an equivalent formulation of the variational Fiedler penalty as a weighted $\text{L}_1$ penalty. We note that the Laplacian quadratic form in the variational Fiedler penalty can be written as $ \textbf{u}^T\textbf{L}\textbf{u} = \sum_{(i,j) \in E} \mathbf{|W|}_{ij} (\textbf{u}(i) - \textbf{u}(j))^2$. This yields the variational objective:
$$ \min_{\mathbf{W}} \mathcal{L}(Y, \mathbf{\hat{f}_W}(X)) + \delta \sum_{(i,j) \in E} \mathbf{|W|}_{ij} (\textbf{u}(i) - \textbf{u}(j))^2$$
We note that both $\mathbf{|W|}_{ij}$ and $(\textbf{u}(i) - \textbf{u}(j))^2$ are non-negative.  This is equivalent to performing $L_1$ penalization on $\mathbf{W}_{ij}$ with weights $(\textbf{u}(i) - \textbf{u}(j))^2$.

There is an immense literature on modified $L_1$ penalties in shallow models \citep{zou2006adaptive, candes2008enhancing}. It is well known that optimizing an objective under (weighted) $L_1$ constraints often yields sparse solutions. This thus connects our Fiedler regularization approach with sparsity induction on the weights of the NN. From an optimization perspective, recall that $\lambda_2(|\mathbf{W}|)$ is a concave function of $|\mathbf{W}|$ and thus a folded concave function of $\mathbf{W}$. One way to optimize this folded concave function is to substitute it with a majorizing surrogate function, in the spirit of majorization-minimization algorithms. It is easy to see from our test vector bound that the weighted $\text{L}_1$ formulation is indeed a majorizer of $\lambda_2(|\mathbf{W}|)$, lending support to our variational approximation from a separate perspective. Similar ideas are explored in \citep{carmichael2021folded}.

In Fiedler regularization, the weights $\mathbf{|W|}_{ij}$ are scaled by a factor of $(\textbf{u}(i) - \textbf{u}(j))^2$, where $\textbf{u}$ is a test vector that approximates $\textbf{v}_2$. From the spectral clustering literature \citep{hagen1992new, donath1972algorithms}, we understand that $\textbf{v}_2$ is very useful for approximating the minimum conductance cut of a graph. The usual heuristic is that one sorts entries of $\textbf{v}_2$ in ascending order, sets a threshold $t$, and groups all vertices $i$ having $\textbf{v}_2 > t$ into one cluster and the rest into another cluster. If the threshold $t$ is chosen optimally, this clustering will be a good approximation to the minimum conductance cut of a graph. As such, the farther apart $\textbf{v}_2(i)$ and $\textbf{v}_2(j)$ are (1) the edge between nodes $i$ and $j$ (if it exists) is likely "less important" for the connectivity structure of the graph, and (2) the more likely that nodes $i$ and $j$ belong to different clusters.

This is also connected to spectral drawing of graphs \citep{Spielman2019Algebraic, hall1970r}, where it can be shown that "nice" planar embeddings/drawings of the graph do not exist if $\lambda_2$ is large. In this sense, Fiedler regularization is forcing the NN to be "more planar" while respecting its connectivity structure.

Hence, with Fiedler regularization, the penalization on $\mathbf{|W|}_{ij}$ is the strongest when this edge is "less important" for the graph's connectivity structure. This would in theory lead to greater sparsity
in edges that have low weights and connect distant vertices belonging to different clusters. Thus, Fiedler regularization sparsifies the NN in a way that respects its connectivity structure.

To this end, an alternative guarantee is the ordering property of Laplacian eigenvalues with respect to a sequence of graphs \citep{godsil2013algebraic}.
\\\\{\bf Proposition 5.1 (Laplacian Eigenvalue Ordering) } {\it Given the graph $G$ with non-negative weights, if we remove an edge $(a,b)$ to obtain $G\setminus (a,b)$, we have that $\lambda_2(G\setminus (a,b)) \leq \lambda_2(G)$.\\\\}
\noindent{\it Proof:}
{
The proof is a simple generalization of \cite{godsil2013algebraic}. Pick $\mathbf{q}_2$ and $\mathbf{r}_2$ to be second eigenvectors of $G$ and $G\setminus(a,b)$ respectively. By the Laplacian quadratic form characterization of eigenvalues, we have $$\lambda_2(G) = \mathbf{q}_2^T\mathbf{L}_G\mathbf{q}_2 = \sum_{(c,d) \in E} \mathbf{|W|}_{cd}(\mathbf{q}_2(c) - \mathbf{q}_2(d))^2$$ $$ \geq [\sum_{(c,d) \in E} \mathbf{|W|}_{cd}(\mathbf{q}_2(c) - \mathbf{q}_2(d))^2] - \mathbf{|W|}_{ab}(\mathbf{q}_2(a) - \mathbf{q}_2(b))^2$$ $$ = \mathbf{q}_2^T\mathbf{L}_{G\setminus (a,b)}\mathbf{q}_2 \geq \mathbf{r}_2^T\mathbf{L}_{G\setminus (a,b)}\mathbf{r}_2 = \lambda_2(G\setminus (a,b))$$
where the first inequality follows from the non-negativity of the weights under consideration and the second inequality follows from the Rayleigh-Ritz variational characterization of eigenvalues. \BlackBox
}
\\\\Hence, by sparsifying edges and reducing edge weights, we are in effect reducing the Fiedler value of the NN. The above ordering property is a special case of the more general Laplacian eigenvalue interlacing property, where $\lambda_2(G\setminus (a,b))$ also admits a corresponding lower bound. For a general treatment, see \cite{godsil2013algebraic}.

We remark that since Fiedler regularization encourages sparsity, during the training process the NN might become disconnected, rendering $\lambda_2$ to become 0. This issue could be easily avoided in practice by dropping one of the disconnected components (e.g. the smaller one) from the Laplacian matrix during training, i.e. remove the Laplacian matrix's rows and columns that correspond to the vertices in the dropped component.

\section{Error Generalization Bounds}
The guarantees that we have provided above focus on the approximation and variational aspects of Fiedler Regularization. In this section, we provide some theoretical guarantees on the generalization error of such an approach. In particular, we use the notion of Rademacher complexity \citep{bartlett2002rademacher} from statistical learning theory to provide finite-sample, uniform generalization error upper bounds for Fiedler regularization. This yields insights into how Fiedler regularization reduces the expressiveness of the learned NN, thus reducing overfitting and improving generalization performance. We focus on the case of supervised classification. We analyze the Laplacian quadratic form/weighted $L_1$ penalization formulation, which includes the exact Fielder value penalization as a special case when the test vector used is exactly $\mathbf{v}_2$. To the best of our knowledge, this is the first Rademacher complexity analysis for a weighted-$L_1$ penalized neural network. The results here might be of independent interest.

\subsection{Rademacher Complexity and Error Bounds}

One key aspect of characterizing the generalization performance of a class of functions/classifiers is via understanding their expressiveness. The notion of empirical Rademacher complexity provides a useful approach for quantifying the expressiveness of a class of functions.
\\\\{\bf Definition 6.1 (Rademacher Complexity) }
{\it Given a sample of $N$ points $\mathbf{z}_1, \cdots, \mathbf{z}_N$ drawn i.i.d. from some probability distribution $\mathcal{P}$ on a set $Z \subseteq \mathbb{R}^d$, as well as a function class $\mathcal{H}$ consisting of functions that map from $Z$ to $\mathbb{R}$, we define the empirical Rademacher complexity of $\mathcal{H}$, denoted $\hat{R}_N(\mathcal{H})$, as:
$$\hat{R}_N(\mathcal{H}) := \mathbb{E}_{\mathcal{R}}\bigg[\sup_{h\in \mathcal{H}} \frac{1}{N}\sum_{i = 1}^N \Omega_i h(\mathbf{z}_i)\Big|\{\mathbf{z}_i\}_{i = 1}^N\bigg]$$
where $\Omega_i$ are independent and identically distributed random variables drawn from the Rademacher distribution $\mathcal{R}$ (i.e. the uniform distribution on $\{-1, 1\}$), and the expectation is taken over the Rademacher random variables $\Omega_i$.
\\\\$\hat{R}_N(\mathcal{H})$ is a random quantity since it conditions on the sample $\mathbf{z}_1, \cdots, \mathbf{z}_N$. The expectation of $\hat{R}_N(\mathcal{H})$ under the product measure $\mathcal{P}^N$ is known as the Rademacher complexity of $\mathcal{H}$, denoted $R_{N, \mathcal{P}}(\mathcal{H})$. In other words,
$$R_{N, \mathcal{P}}(\mathcal{H}) := E_{\mathcal{P}^N}[\hat{R}_N(\mathcal{H})]$$
}\\
On an intuitive level, the Rademacher complexity of the function class $\mathcal{H}$ characterizes how "expressive" $\mathcal{H}$ is, in terms of the ability of $\mathcal{H}$ to match a random sign pattern (i.e. the Rademacher random variables). The larger the Rademacher complexity, the more expressive the function class, and this generally indicates poorer generalization performance due to overfitting. Note that the expectation in the definition of the Rademacher complexity depends on $\mathcal{P}$, which in many problems is complicated, high-dimensional and/or unknown. The notion of empirical Rademacher complexity bypasses this issue by conditioning on the sample.

The Rademacher complexity is a useful notion in part because it naturally occurs in many bounds of interest in empirical process theory and statistical learning theory. For the case of supervised classification where $Z = (X, Y)$, one can choose $\mathcal{H} := \{(x, y) \mapsto \mathcal{L}(f(x), y): f \in \mathcal{F}\}$, where $\mathcal{L}$ is a fixed loss function, and $\mathcal{F}$ is the class of classifiers under consideration. In the case of classification, the function $\mathcal{L}$ is commonly chosen to be a surrogate loss, such as the hinge loss or the logistic loss. These commonly used loss functions are Lipschitz continuous with Lipschitz constant 1. The following theorem is then useful.
\\\\{\bf Theorem 6.2 (Generalization Bounds via Rademacher Complexity) }
{\it Let $\mathcal{H} := \{(x, y) \mapsto \mathcal{L}(f(x), y): f \in \mathcal{F}\}$, where $\mathcal{L}$ is a 1-Lipschitz continuous loss function. Then for any $\delta \in (0,1)$, with probability at least $1-\delta$ under the product measure $\mathcal{P}^N$, the following inequality holds:
$$\sup_{h \in \mathcal{H}} \left[ \frac{1}{N}\sum_{i = 1}^N h(\mathbf{z}_i) - \mathbb{E}_{\mathcal{P}}(h(\mathbf{z})) \right] \leq  2\hat{R}_N(\mathcal{H}) + O\bigg(\sqrt{\frac{\log(1/\delta)}{N}}\bigg),$$
where each $\mathbf{z}_i$ is distributed according to $\mathcal{P}$ independently. }\\

The proof of the above theorem could be found in many sources and textbooks, including \cite{bartlett2002rademacher, shalev2014understanding, bousquet2003introduction}.  The above probabilistic bound is uniform and finite-sample. When applied to the learning problem, the quantity $\mathbb{E}_{\mathcal{P}}(h(\mathbf{z}))$ corresponds to the "true" error whereas the quantity $\frac{1}{N}\sum_{i = 1}^N h(\mathbf{z}_i)$ corresponds to the error obtained on the training set. Thus theorem 6.2 provides a way to control the generalization error, in a uniform, finite-sample and distribution-agnostic way, which is particularly suited to analyzing modern machine learning models. The asymptotic notation $O(\cdot)$ is used here to hide constant factors that are very small. Explicit forms of the inequality with the constant factors shown are described in the sources cited above.

\subsection{Rademacher Complexity Bounds for Neural Networks}

Armed with the above theorem, our focus turns to upper-bounding the empirical Rademacher complexity $\hat{R}_N(\mathcal{H})$ for feedforward neural networks penalized by Fiedler regularization. This in turn shows us how Fiedler regularization affects generalization error bounds of neural networks.
Our strategy is to first upper bound the Rademacher complexity $\hat{R}_N(\mathcal{F})$ of the class of feedforward neural networks classifiers under consideration. We then use the results on $\hat{R}_N(\mathcal{F})$ to derive upper bounds on $\hat{R}_N(\mathcal{H})$ by applying the loss function.

To gain intuition to our approach, recall that we operate in the following setting: fix the architecture of a neural network, setting $\sigma$ to be the activation function and $\Lambda$ to be the number of layers. Then the feedforward neural network under our consideration would have the following form:
\begin{equation}\label{eq:nn}
   f(\mathbf{x}) := \mathbf{W}^{(\Lambda)} \sigma(\mathbf{W}^{(\Lambda-1)}(\sigma(\cdots \sigma(\mathbf{W}^{(1)}\mathbf{x})\cdots)))
\end{equation}
where the activation function $\sigma$ is applied component-wise. Note that biases are absorbed into the inputs $\mathbf{x}$ without loss of generality. Also note that $\mathbf{W}^{(\Lambda)}$ denotes that matrix encoding the linear transformation that maps nodes from the $l-1$\textsuperscript{th} layer to nodes in the $l$\textsuperscript{th} layer.

To prove Rademacher complexity bounds for neural networks, we first list several useful lemmas. Given the compositional nature of equation \ref{eq:nn}, the first lemma below helps control the Rademacher complexity of function classes under Lipschitz compositions.
\\\\{\bf Lemma 6.3 (Talagrand's Contraction Lemma) \citep{ledoux2013probability}}
{\it Let $\phi_1, \cdots \phi_N$ be $\gamma$-Lipschitz functions mapping from $\mathbb{R} \to \mathbb{R}$ for some $\gamma > 0$. Then
$$\hat{R}_N((\phi_1, \cdots, \phi_N) \circ \mathcal{H}) := \mathbb{E}_{\mathcal{R}}\bigg[\sup_{h\in \mathcal{H}} \frac{1}{N}\sum_{i = 1}^N \Omega_i \phi_i(h(\mathbf{z}_i))\Big|\{\mathbf{z}_i\}_{i = 1}^N\bigg] $$ $$\leq \gamma \mathbb{E}_{\mathcal{R}}\bigg[\sup_{h\in \mathcal{H}} \frac{1}{N}\sum_{i = 1}^N \Omega_i h(\mathbf{z}_i)\Big|\{\mathbf{z}_i\}_{i = 1}^N\bigg] = \gamma \hat{R}_N(\mathcal{H})$$
where the $\circ$ notation denotes function composition. }\\\\
In other words, we can use the (empirical) Rademacher complexity of $\mathcal{H}$ to control its (empirical) Rademacher complexity after an element-wise composition with $\gamma$-Lipschitz functions. Talagrand's contraction lemma provides a useful way to analytically control the expressivity of functions/classifiers that exhibit compositional representations.

Feedforward NNs naturally admit compositional descriptions, with the original input data undergoing interleaving affine transformations and activations. Virtually all activation functions that are used in practice are Lipschitz continuous, with popular choices such as tanh, ReLU, etc being 1-Lipschitz. Popular operators, most notably the convolutional operator commonly used in compution settings, are also Lipschitz. Talagrand's contraction lemma thus allow us to study broad classes of feedforward neural networks.

Linear transformations also form a crucial part of neural networks. The following lemma gives us control over Rademacher complexity of linear transformations under weights with bounded $L_1$ norm.
\\\\{\bf Lemma 6.4 (Rademacher Complexity of a Linear Class)}
{\it  Given an i.i.d. sample $\{\mathbf{x}_i\in \mathbb{R}^d\}_{i = 1}^N$ and a function class $\mathcal{F}_{linear}^B:= \{\mathbf{x} \mapsto \langle \mathbf{w}, \mathbf{x} \rangle \;| \; \mathbf{w} \in \mathbb{R}^d,\; ||\mathbf{w}||_1 \leq B \}$, suppose that $||\mathbf{x}_i||_\infty \leq C$ for all $\mathbf{x}_i$ in the dataset, we have $$\hat{R}_N(\mathcal{F}_{linear}^B) \leq BC\sqrt{\frac{2\log(2d)}{N}}$$
}\\
This is a standard result whose proof is based on a lemma of \cite{massart2000some}. A complete proof can be found in many references, for example section 26.2 of \cite{shalev2014understanding}.

The assumption of bounded $\text{L}_1$ norm of the weights $\mathbf{w}$ is appropriate for our context of studying neural network sparsity, and the assumption of uniformly bounded $\text{L}_\infty$ norm on the data points in the dataset is usually going to be satisfied in practice.

A small modification of the above lemma would lend itself to the scenario of a weighted $L_1$ scenario that is applicable to the Fiedler regularization setting:
\\\\{\bf Corollary 6.5 (Rademacher Complexity of a Linear Class under Weighted $L_1$ Constraints)}
{\it  Given an i.i.d. sample $\{\mathbf{x}_i\in \mathbb{R}^d\}_{i = 1}^N$ and a function class $\mathcal{F}_1:= \{\mathbf{x} \mapsto \langle \mathbf{w}, \mathbf{x} \rangle \;| \; \mathbf{w} \in \mathbb{R}^d,\; \sum_{i = 1}^d \mathbf{c}_i|\mathbf{w}_i| \leq B \}$, where $\mathbf{c} \geq \mathbf{0}$ is a fixed $d$-dimensional weighting vector, suppose that $||\mathbf{x}_i||_\infty \leq C$ for all $\mathbf{x}_i$ in the dataset, we have $$\hat{R}_N(\mathcal{F}_1) \leq \frac{B}{\min(\mathbf{c})}C\sqrt{\frac{2\log(2d)}{N}}$$
where we use $\min(\mathbf{c})$ to denote the minimum entry in the vector $\mathbf{c}$.
}\\\\
{\it Proof:} {  Note that $\min(\mathbf{c}) \sum_{i = 1}^d |\mathbf{w}_i| \leq \sum_{i = 1}^d \mathbf{c}_i|\mathbf{w}_i| \leq B$. This implies $||\mathbf{w}||_1 \leq \frac{B}{\min(\mathbf{c})}$. Note that this implies $\mathcal{F}_1 \subseteq \mathcal{F}_{linear}^{\frac{B}{\min(\mathbf{c})}}$. Invoking Lemma 6.4, we obtain the desired upper bound $\hat{R}_N(\mathcal{F}_1)\leq \hat{R}_N(\mathcal{F}_{linear}^{\frac{B}{\min(\mathbf{c})}}) \leq \frac{B}{\min(\mathbf{c})}C\sqrt{\frac{2\log(2d)}{N}}$. \BlackBox \\
}\\
Recall that by the Langrangian dual formulation in optimization, an $L_1$ penalty added to the objective corresponds to imposing a $L_1$ constraint on the domain on the objective. Analogously, in the case of Fiedler regularization, since a weighted $L_1$ penalty is added to the objective, it is equivalent to imposing a weighted $L_1$ constraint on the weights.

We now have the tools to control the Rademacher complexity a neural network layer by layer. Let $d_0 = d$ be the dimension of the input data, and $d_l$ be the width of layer $l$ in the neural network. Let $\mathbf{c}^0, \mathbf{c}^1, \cdots, \mathbf{c}^{l-1}$ be non-negative weighting vectors of dimensions $d_0, d_1, \cdots, d_{l-1}$ respectively. Let $B_0, B_1, \cdots, B_{l-1}$ be non-negative constants. Let the base class be $\mathcal{F}_1:= \{\mathbf{x} \mapsto \langle \mathbf{w}, \mathbf{x} \rangle \;| \; \mathbf{w} \in \mathbb{R}^d,\; \sum_{i = 1}^d \mathbf{c}^0_i|\mathbf{w}_i| \leq B_0 \}$.
Recursively define $\mathcal{F}_l$ for $l$ running through $2$ to $\Lambda$ as
$$\mathcal{F}_l := \{ \mathbf{x} \mapsto \sum_{i = 1}^{d_{l-1}} \mathbf{w}_i \sigma(f_i(\mathbf{x})) \mid f_i \in \mathcal{F}_{l - 1}, \sum_{i = 1}^{d_{l-1}} \mathbf{c}^{l-1}_i|\mathbf{w}_i| \leq B_{l-1} \}$$
Currently, we leave $\mathbf{c}^0, \mathbf{c}^1, \cdots, \mathbf{c}^{l-1}$ as generic non-negative weight vectors to derive a general result, and later we choose them to correspond to the case of Fiedler regularization specifically.
\\\\
We can now state our main result:
\\\\{\bf Theorem 6.6 (Rademacher Complexity Bound for Neural Network under Weighted $L_1$ Penalty)}
{\it Let $\mathcal{F}_\Lambda$ denote the class of $\Lambda$-layer neural networks with a fixed feedforward architecture and $\gamma$-Lipschitz activations applied element-wise, where the weight matrix for the l\textsuperscript{th} layer is denoted $\mathbf{W}^{(l)}$. Assume the input features $\{\mathbf{x}_i\in \mathbb{R}^d\}_{i = 1}^N $ satisfy $||\mathbf{x}_i||_\infty \leq C$ for all $\mathbf{x}_i$. Also assume that for every layer $l$, the $\mathbf{c}^{l - 1 }$-weighted $\text{L}_1$ norm of every row of $\mathbf{W}^{(l)}$ is upper bounded by $B_{l -1}$. Then we have
$$\hat{R}_N(\mathcal{F}_\Lambda) \leq (2\gamma)^{\Lambda-1} ( \prod_{l =1}^{\Lambda} \frac{B_{l-1}}{\min(\mathbf{c}^{l-1})}) C \sqrt{\frac{2\log(2d)}{N}}$$}
\\{\it Proof:} { It is easy to see that the rows of the weight matrices $\mathbf{W}^{(l)}$ corresponds precisely to the weights $\mathbf{w}$ of the linear transformation applied in $\mathcal{F}_l$. This means that the assumption where each row of $\mathbf{W}^{(l)}$ has $\mathbf{c}^{l-1}$-weighted $L_1$ norm at most $B_{l -1}$ corresponds exactly to the constraints in $\mathcal{F}_l$. Hence we can proceed with the analysis via the function classes $\mathcal{F}_l$.
\\\\We first note that \begin{equation} \label{eq:one_layer}
    \hat{R}_N(\mathcal{F}_\Lambda) \leq 2\gamma \frac{B_{\Lambda - 1}}{\min(\mathbf{c}^{\Lambda - 1})}\hat{R}_N(\mathcal{F}_{\Lambda-1})
\end{equation} To see this, use the definition
$$\hat{R}_N(\mathcal{F}_\Lambda) = \mathbb{E}_{\mathcal{R}}\bigg[\sup_{f\in \mathcal{F}_\Lambda} \frac{1}{N}\sum_{i = 1}^N \Omega_i f(\mathbf{z}_i)\Big|\{\mathbf{z}_i\}_{i = 1}^N\bigg]$$
$$ = \mathbb{E}_{\mathcal{R}}\bigg[\sup_{f_j\in \mathcal{F}_{\Lambda-1}, \sum_{j = 1}^{d_{\Lambda - 1}}\mathbf{c}^{\Lambda - 1}_j\mathbf{w}_j \leq B_{\Lambda - 1}} \frac{1}{N}\sum_{i = 1}^N \Omega_i \sum_{j= 1}^{d_{\Lambda - 1}} \mathbf{w}_j \sigma(f_j(\mathbf{z}_i))\Big|\{\mathbf{z}_i\}_{i = 1}^N\bigg]$$
$$= \mathbb{E}_{\mathcal{R}}\bigg[\sup_{f_j\in \mathcal{F}_{\Lambda-1}, \sum_{j = 1}^{d_{\Lambda - 1}}\mathbf{c}^{\Lambda - 1}_j\mathbf{w}_j \leq B_{\Lambda - 1}} \sum_{j= 1}^{d_{\Lambda - 1}}\mathbf{w}_j\frac{1}{N}\sum_{i = 1}^N \Omega_i   \sigma(f_j(\mathbf{z}_i))\Big|\{\mathbf{z}_i\}_{i = 1}^N\bigg]$$
Apply Cauchy-Schwartz, get:
$$\leq \mathbb{E}_{\mathcal{R}}\bigg[\sup_{f_j\in \mathcal{F}_{\Lambda-1}, \sum_{j = 1}^{d_{\Lambda - 1}}\mathbf{c}^{\Lambda - 1}_j\mathbf{w}_j \leq B_{\Lambda - 1}} ||\mathbf{w}||_1 \max_j \left|\frac{1}{N}\sum_{i = 1}^N \Omega_i   \sigma(f_j(\mathbf{z}_i))\right|\Big|\{\mathbf{z}_i\}_{i = 1}^N\bigg]$$
Using the constraint $ \sum_{j = 1}^{d_{\Lambda - 1}}\mathbf{c}^{\Lambda - 1}_j\mathbf{w}_j \leq B_{\Lambda - 1}$, which implies $||\mathbf{w}||_1 \leq \frac{B_{\Lambda - 1}}{\min(\mathbf{c}^{\Lambda - 1})}$, we have:
$$\leq  \frac{B_{\Lambda - 1}}{\min(\mathbf{c}^{\Lambda - 1})} \mathbb{E}_{\mathcal{R}}\bigg[\sup_{f\in \mathcal{F}_{\Lambda-1}} \left|\frac{1}{N}\sum_{i = 1}^N \Omega_i   \sigma(f(\mathbf{z}_i))\right|\Big|\{\mathbf{z}_i\}_{i = 1}^N\bigg]$$
$$\leq  2\frac{B_{\Lambda - 1}}{\min(\mathbf{c}^{\Lambda - 1})} \mathbb{E}_{\mathcal{R}}\bigg[\sup_{f\in \mathcal{F}_{\Lambda-1}} \frac{1}{N}\sum_{i = 1}^N \Omega_i   \sigma(f(\mathbf{z}_i))\Big|\{\mathbf{z}_i\}_{i = 1}^N\bigg]$$
Applying lemma 6.3 (Talagrand's contraction lemma), get:
$$\leq  2\gamma\frac{B_{\Lambda - 1}}{\min(\mathbf{c}^{\Lambda - 1})} \mathbb{E}_{\mathcal{R}}\bigg[\sup_{f\in \mathcal{F}_{\Lambda-1}} \frac{1}{N}\sum_{i = 1}^N \Omega_i f(\mathbf{z}_i)\Big|\{\mathbf{z}_i\}_{i = 1}^N\bigg]$$
$$ = 2\gamma\frac{B_{\Lambda - 1}}{\min(\mathbf{c}^{\Lambda - 1})} \hat{R}_N(\mathcal{F}_{\Lambda - 1})$$
This shows the claim \ref{eq:one_layer}.
\\\\Applying the inequality \ref{eq:one_layer} recursively $\Lambda -1$ times, we obtain
$$  \hat{R}_N(\mathcal{F}_\Lambda) \leq 2^{\Lambda-1} \gamma^{\Lambda -1} \left(\prod_{l = 2}^\Lambda \frac{B_{l - 1}}{\min(\mathbf{c}^{l -1})}\right)\hat{R}_N(\mathcal{F}_{1}) $$
Applying corollary 6.5, get:
$$  \hat{R}_N(\mathcal{F}_\Lambda) \leq 2^{\Lambda-1} \gamma^{\Lambda -1} \left(\prod_{l = 2}^\Lambda \frac{B_{l - 1}}{\min(\mathbf{c}^{l -1})}\right)  \frac{B_0}{\min(\mathbf{c}^0)}C\sqrt{\frac{2\log(2d)}{N}}$$
$$ =   (2\gamma)^{\Lambda -1} \prod_{l = 1}^\Lambda \frac{B_{l - 1}}{\min(\mathbf{c}^{l -1})}  C\sqrt{\frac{2\log(2d)}{N}}$$ This shows the desired result. \BlackBox
}\\\\
The above result allows us to control the Rademacher complexity of a neural network under a weighted $L_1$ penalty in a layer-wise fashion. In the context of Fiedler regularization, the penalty applied is $\sum_{(a, b) \in E} (\mathbf{u}(a) - \mathbf{u}(b))^2 |\mathbf{W}_{ab}|$, where we recall that the $\mathbf{W}$ without superscript denotes the weighted adjacency of the underlying graph of the entire neural network, and $\mathbf{u}$ being a fixed test vector (which can be chosen as the second Laplacian eigenvector $\mathbf{v}_2$ for exact results). We can rewrite this in matrix form by defining a $n \times n$ matrix $\mathbf{U}$ as $\mathbf{U}_{ab} = (\mathbf{u}(a) - \mathbf{u}(b))^2$. We thus obtain an equivalent expression of the penalty term as
$\sum_{(a, b) \in E}  \mathbf{U}_{ab}|\mathbf{W}_{ab}|$.
\\\\This penalty is separable, and we can split it by layers. Denote the nodes in layer $l$ by $V_l$ and the edges that connects nodes from $V_{l-1}$ to $V_l$ as $E_l$. We obtain the equivalent expression for the penalty term $\sum_{l = 1}^{\Lambda} \sum_{(a,b) \in E_l} \mathbf{U}_{ab} |\mathbf{W}^{(l)}_{ab}|$. This now has a weighted $L_1$ form that is similar to the layer-wise weighted $L_1$ constraints used in the function classes $\mathcal{F}_l$. Thus, the weighting vectors $\mathbf{c}^0, \mathbf{c}^1, \cdots, \mathbf{c}^{\Lambda - 1}$ used in theorem 6.6 above can now be chosen accordingly.
\\\\Recall that $$\mathcal{F}_l := \{ \mathbf{x} \mapsto \sum_{i = 1}^{d_{l-1}} \mathbf{w}_i \sigma(f_i(\mathbf{x})) \mid f_i \in \mathcal{F}_{l - 1}, \sum_{i = 1}^{d_{l-1}} \mathbf{c}^{l-1}_i|\mathbf{w}_i| \leq B_{l-1} \}$$ The weights $\mathbf{w}$ in $\mathcal{F}_l$ map the nodes in $V_{l-1}$ to a single node in layer $l$, and corresponds to a row of the matrix $\mathbf{W^{(l)}}$.
Thus one way to select $\mathbf{c}^{l-1}$ is to pick its $b$\textsuperscript{th} entry, where $b$ indexes the nodes in $V_{l-1}$, as the maximum entry inside the $b$\textsuperscript{th} column of $\mathbf{U}$. Due to the feedforward nature of the network (without any skip connections), this corresponds to finding the maximum of $\mathbf{U}$ over the edges that connects node $b$ in $V_{l-1}$ to all the nodes in $V_l$, i.e. $\mathbf{c}^{l-1}(b) = \max_{\{a \mid (a,b) \in E_l\}} \mathbf{U}_{a,b}$. Selecting the max entry along the $b$\textsuperscript{th} column ensures that the resulting function class $\mathcal{F}_\Lambda$ is large enough to include the Fiedler regularized neural network.
\\\\We then have the following corollary:
\\\\{\bf Corollary 6.7 (Rademacher Complexity Bound for Fiedler Regularized Neural Network)}
{\it  Assume the same conditions of Theorem 6.6. In addition, fix a test vector $\mathbf{u}$ for Fiedler regularization, which specfies a matrix $\mathbf{U}$. For each $l$ from $1$ to $\Lambda$, select weighting vectors $\mathbf{c}^{l-1}$ according to $\mathbf{c}^{l-1}(b) = \max_{\{a \mid (a,b) \in E_l\}} \mathbf{U}_{a,b}$, where $b$ indexes $V_{l - 1}$. The empirical Rademacher complexity of the resulting function class is upper bounded as follows:
$$\hat{R}_N(\mathcal{F}_\Lambda) \leq (2\gamma)^{\Lambda-1} ( \prod_{l =1}^{\Lambda} \frac{B_l}{\min_{\{b \in V_{l-1}\}}(\max_{\{a \mid (a,b) \in E_l\}}\mathbf{U}_{a,b} )}) C \sqrt{\frac{2\log(2d)}{N}}$$
}\\\\{\it Proof:} {Simply set the vectors $\mathbf{c}^{l-1}$ according to $\mathbf{c}^{l-1}(b) $ $= \max_{\{a \mid (a,b) \in E_l\}} \mathbf{U}_{a,b}$ and apply Theorem 6.6.  \BlackBox}\\\\
There are more recent results \citep{golowich2018size} that use a more refined Rademacher analysis to remove the exponential dependency on depth from the $(2\gamma)^{\Lambda-1}$ term. However, building on such results to obtain more refined bounds is a future direction, and does not affect the interpretation/derivations for Fielder regularization, which controls the norm of the weight matrices.

The above bound in corollary 6.5 yields insight into how Fiedler regularization works. A larger value for $\mathbf{U}_{ab} = (\mathbf{u}(a) - \mathbf{u}(b))^2$ implies that the edge $(a, b)$ is "less important" for connectivity. Thus the max operation $\max_{\{a \mid (a,b) \in E_l\}}\mathbf{U}_{a,b})$ can be interpreted as looking for the least important edge that node $b$ is associated with. The subsequent $\min_{\{b \in V_{l-1}\}}$ operation finds the node in layer $l-1$ whose least important edge has the best connectivity. It can be thought of as a way of characterizing the connectivity bottleneck inside a layer in a minimax/worst-case manner. The higher this value, the stronger the penalization, and hence the smaller the Rademacher complexity.

We can combine theorem 6.2 and corollary 6.7 to obtain the generalization error bound on a Fiedler regularized neural network.
\\\\{\bf Corollary 6.8 (Generalization Error Bound for Fiedler Regularized Neural Network)}
{\it  Let $\mathcal{H} := \{(x, y) \mapsto \mathcal{L}(f(x), y): f \in \mathcal{F}_\Lambda \}$, where $\mathcal{L}$ is a 1-Lipschitz continuous loss function. Then for any $\delta \in (0,1)$, with probability at least $1-\delta$ under the product measure $\mathcal{P}^N$, the following inequality holds:
$$\sup_{h \in \mathcal{H}} \left[ \frac{1}{N}\sum_{i = 1}^N h(\mathbf{z}_i) - \mathbb{E}_{\mathcal{P}}(h(\mathbf{z})) \right] $$ $$\leq  2(2\gamma)^{\Lambda-1} ( \prod_{l =1}^{\Lambda} \frac{B_l}{\min_{\{b \in V_{l-1}\}}(\max_{\{a \mid (a,b) \in E_l\}}\mathbf{U}_{a,b} )}) C \sqrt{\frac{2\log(2d)}{N}} + O\bigg(\sqrt{\frac{\log(1/\delta)}{N}}\bigg),$$
where each $\mathbf{z}_i$ is distributed according to $\mathcal{P}$ independently.
}\\
\\{\it Proof:} { Apply Talagrand's contraction lemma to obtain $\hat{R}_N(\mathcal{H}) \leq 1 \times \hat{R}_N(\mathcal{F}_\Lambda)$, and then apply theorem 6.2. \BlackBox}

\section{Supervised Classification: Experiments and Results}
Deep/multilayer feedforward NNs are useful in many classification problems, ranging from popular image recognition tasks to scientific and biomedical problems such as classifying diseases.  We examine the performance of Fiedler regularization on the standard benchmark image classification datasets MNIST and CIFAR10. We also tested Fiedler regularization on the TCGA RNA-Seq PANCAN tumor classification dataset \citep{weinstein2013cancer} from the UCI Machine Learning Repository. We compare the performance of several standard regularization approaches for NNs on these datasets, including Dropout, $\text{L}_1$ regularization and weight decay. Extension of such experiments to other classification tasks is straightforward.

The purpose of the experiments below is not to use the deepest NNs, the latest architectures or the most optimized hyperparameters. Nor is the purpose to show the supremacy of NNs versus other classification methods like random forests or logistic regression. Rather, we attempt to compare the efficacy of Fiedler regularization against other NN regularization techniques as a proof of concept. Extensions to more complicated and general network architectures are explored in the discussion section.

For all our experiments, we consider 5-layer feedforward NNs with ReLU activations and fully connected layers. We used PyTorch 1.4 and Python 3.6 for all experiments. We adopt stochastic gradient descent with a momentum of 0.9 for optimization and a learning rate of 0.001. To select the dropping probability for dropout, as well as the regularization hyperparameter for $\text{L}_1$, Fiedler regularization and weight decay, we performed a very rough grid search on a small validation dataset. The dropout probability is selected to be 0.5 for all layers, and the regularization hyperparameters for $\text{L}_1$, Fiedler regularization and weight decay are 0.001, 0.01 and 0.01 respectively. All models in the experiments were trained under the cross-entropy loss. Each experiment was run 5 times, with the median and the standard deviation of the performances reported. All experiments were run on a Unix machine with an Intel Core i7 processor. The code used for the experiments can be found at the first author's Github repository (\href{https://github.com/edrictam/FiedlerRegularization}{https://github.com/edrictam/FiedlerRegularization}).

\subsection{MNIST}
\textbf{Dataset and setup}. MNIST is a standard handwriting recognition dataset that consists of 60,000 $28 \times 28$ training images of individual hand-written digits and $10,000$ testing images. We picked the hidden layers of our NN to be 500 units wide. We used a batch size of 100 and the networks were trained with 10 epochs.
\\\textbf{Results}. The results for MNIST are displayed in Table 1. For the MNIST dataset, we obtained very good accuracies for all the methods, with Fiedler regularization standing out, followed by weight decay and dropout. The high accuracies obtained for the MNIST dataset with feedforward NNs are consistent with results from previous studies \citep{srivastava2014dropout}. Fiedler regularization showed gains over its competitors.

\begin{table}[t]
\caption{Classification accuracies for MNIST under various regularization schemes (units in percentages)}
\label{sample-table-1}
\vskip 0.15in
\begin{center}
\begin{small}
\begin{sc}
\begin{tabular}{lccr}
\toprule
Regularization & Training & Testing  \\
\midrule
$\text{L}_1$   & 90.32  $\pm$ 0.17& 90.25$\pm$ 0.35\\
Weight Decay & 95.12$\pm$ 0.06& 94.98$\pm$ 0.07\\
Dropout & 94.52$\pm$ 0.07 &94.5$\pm$ 0.18 \\
Fiedler & 96.54$\pm$ 0.08 &96.1$\pm$ 0.12 \\
\bottomrule
\end{tabular}
\end{sc}
\end{small}
\end{center}
\vskip -0.1in
\end{table}

\subsection{CIFAR10}
\textbf{Dataset and setup}.
CIFAR10 is a benchmark object recognition dataset that consists of $32\times32\times 3$ down-sampled RGB color
images of 10 different object classes. There are 50,000 training images and 10,000 test images in the dataset. We picked the hidden layers of our NN to be 500 units wide. We used a batch size of 100 and the networks were trained with 10 epochs.
\\\textbf{Results}. The results for CIFAR10 are displayed in Table 2. While this is a more difficult image classification task than MNIST, the ordering of performances among the regularization methods is similar. Fiedler regularization performed the best, followed by dropout, weight decay and $\text{L}_1$ regularization.

\begin{table}[t]
\caption{Classification accuracies for CIFAR10 under various regularization schemes (units in percentages)}
\label{sample-table-2}
\vskip 0.15in
\begin{center}
\begin{small}
\begin{sc}
\begin{tabular}{lccr}
\toprule
Regularization & Training & Testing \\
\midrule
$\text{L}_1$   & 28.52$\pm$ 0.9& 28.88$\pm$ 1.11\\
Weight Decay  & 52.93 $\pm$ 0.17& 50.55$\pm$ 0.39\\
Dropout & 46.61$\pm$ 0.35& 44.63$\pm$ 0.39 \\
Fiedler & 57.99 $\pm$ 0.13 & 52.26$\pm$ 0.27 \\
\bottomrule
\end{tabular}
\end{sc}
\end{small}
\end{center}
\vskip -0.1in
\end{table}

\subsection{TCGA Cancer Classification}
\textbf{Dataset and setup}. The TCGA Pan-Cancer tumor classification dataset consists of RNA-sequencing as well as cancer classification results for 800 subjects. The input features are 20531-dimensional vectors of gene expression levels, whereas the outputs are tumor classification labels (there are 5 different tumor types under consideration). We used 600 subjects for training and 200 for testing. Due to the highly over-parametrized nature of this classification task, We picked the width of the hidden layers to be narrower, at 50 units. We used a batch size of 10 and the networks were trained with 5 epochs.
\\\textbf{Results}
The results for the TCGA tumor classification experiment are displayed in Table 3. Fiedler regularization and $\text{L}_1$ had similarly high performances, followed by weight decay. It is interesting that Dropout achieved a relatively low accuracy, slightly better than chance.  Since this dataset is relatively small (the testing set has only 200 data points and training set 600), the standard deviation of the accuracies are higher. $\text{L}_1$ and Fiedler regularization, which explicitly induce sparsity, performed the best.
\begin{table}[t]
\caption{Classification accuracies for TCGA under various regularization schemes (units in percentages)}
\label{sample-table-3}
\vskip 0.15in
\begin{center}
\begin{small}
\begin{sc}
\begin{tabular}{lccr}
\toprule
Regularization & Training & Testing  \\
\midrule
$\text{L}_1$   & 91.5  $\pm$ 13.73& 94.53$\pm$ 12.46\\
Weight Decay & 57$\pm$ 22.42& 60.2$\pm$ 20.94\\
Dropout & 25.33$\pm$ 5.9 &23.88$\pm$ 7.36 \\
Fiedler & 93.33$\pm$ 20.31 &90.55$\pm$ 20.73 \\
\bottomrule
\end{tabular}
\end{sc}
\end{small}
\end{center}
\vskip -0.1in
\end{table}

\subsection{Analysis of Results}

We note that both MNIST and CIFAR10 have more training samples than the dimension of their features. The results from MNIST and CIFAR10 largely agree with each other and confirm the efficacy of Fiedler regularization. Under this setting, other regularization methods like dropout and weight decay also exhibited decent performance.

On the other hand, in the TCGA dataset, where the dimension of the input features (20531) is much higher than the number of training samples (600), $\text{L}_1$ and Fiedler regularization, which explicitly induce sparsity, performed substantially better than dropout and weight decay. An inspection of the TCGA dataset suggests that many of the gene expression levels in the input features are 0 (suggesting non-expression of genes), which likely implies that many of the weights in the network, particularly at the input layer, are not essential.  It is thus not surprising that regularization methods that explicitly induce sparsity perform better in these "large p, small n" scenarios, often found in biomedical applications.

We remark that Fiedler regularization enjoys practical running speeds that are fast, generally comparable to (but slightly slower than) that of most commonly used regularization schemes such as $\text{L}_1$ and dropout. The running time of Fiedler regularization could likely be improved with certain implementation-level optimizations to speed up the software. Performances would also likely improve if more refined grid searches or more sophisticated tuning parameter selection methods like Bayesian optimization are adopted.

\section{Discussion}
The above experiments demonstrated several points of interest. The poorer performance of $\text{L}_1$ regularization in MNIST and CIFAR10 stands in sharp contrast to the much higher performance of Fiedler regularization, a weighted $\text{L}_1$ penalty. It is generally acknowledged that $\text{L}_1$ regularization does not enjoy good empirical performance in deep learning models. The precise reason why this happens is not exactly known.  Previous studies have adopted a group-lasso formulation for regularization of deep NNs and have obtained good performance \citep{scardapane2017group}. These results suggest that modifications of $\text{L}_1$ through weighting or other similar schemes can often drastically improve empirical performance.

We have tracked the algebraic connectivity of the NNs during the training process in our experiments. In general, without any regularization, the NNs tend to become more connected during training, i.e. their Fiedler value increases. In the Fiedler regularization case, the connectivity is penalized and therefore decreases during training in a very gradual manner. Interestingly, in the $\text{L}_1$ case, the algebraic connectivity of the NN can decrease very quickly during training, often leading to disconnection of the network very early in the training process. This is likely related to $\text{L}_1$ regularization's uniform penalization of all weights. It is therefore difficult to choose the regularization hyperparameter for $\text{L}_1$: if it is too small, sparsity induction might occur too slowly and we would under-regularize; if it is too big, we risk over-penalizing certain weights and lowering  the model's accuracy. One advantage of a weighted scheme such as Fiedler penalization thus lies in its ability to adaptively penalize different weights during training.

While we only considered relatively simple feedforward, fully connected neural architectures, potential extensions to more sophisticated structures are straightforward. Many convolutional NNs contain fully connected layers after the initial convolutional layers. One could easily extend Fiedler regularization to this case. In the context of ResNets, where there are skip connections, the spectral graph properties we have utilized still hold, and hence Fiedler regularization could be directly applied. In the non-classification setting, autoencoders exhibit very natural graph structures in the form of a bottleneck, and it is an open direction to study graph connectivity based penalizations, such as Fiedler regularization, in that setting. The spectral graph theory setup adopted in this paper generally holds for any undirected graph. An open direction is to establish appropriate spectral graph theory for regularization of directed graphs, which would be useful in training recurrent NNs.

While Fiedler regularization leads to sparsely connected NNs in theory, in practice it often takes a higher penalty value or longer training time to achieve sparsity with Fiedler regularization. This might be due in part to the optimization method chosen. It is known that generally SGD does not efficiently induce sparsity in $\text{L}_1$ penalized models, and certain truncated gradient methods \citep{langford2009sparse} might prove more effective in this setting.

We remark that while Fiedler regularization emphasizes regularizing based on graphical/connectivity structure, global penalization approaches such as Dropout, L2 etc could still prove useful. One could combine the two regularization methodologies to achieve simultaneous regularization.

There are many important statistical and machine learning models that can be cast as special cases of shallow neural networks, e.g. factor analysis, linear regression etc. One natural direction of further investigation is to apply Fiedler regularization to these models.

There are intimate connections between the graph cut/edge expansion perspective considered above, and the geometric notion of isoperimetry. The Laplacian matrix considered above could be viewed as a discretized analog of the Laplace-Beltrami operator. An open direction is to investigate continuous relaxations of the Fiedler regularization approach.

On the generalization error bound analysis, there is an independent line of work \citep{neyshabur2017pac} that utilizes the spectral norm of the neural network's weight matrices to provide margin bounds on the generalization error. Their focus is on utilizing the spectral norms of the weight matrices for the layers, whereas our focus is on leveraging the global structure via the spectrum of the Laplacian of the NN's underlying graph.

Lastly, we have adopted a version of spectral graph theory that considers the un-normalized (combinatorial) Laplacian $\mathbf{L} = \mathbf{D}-\mathbf{|W|}$ as well as the edge expansion of the graph. A similar theory for regularization could be developed for the normalized Laplacian $\mathbf{L}' = \mathbf{I} - \mathbf{D}^{-\frac{1}{2}}\mathbf{|W|}\mathbf{D}^{-\frac{1}{2}}$ and the conductance of the graph, after appropriately accounting for the total scale of the NN. While the combinatorial Laplacian that we considered is related to the notion of RatioCuts, the normalized Laplacian is associated with the notion of NCuts. Both notions could in theory be used for reducing connectivity/co-adaptation of the NN.

\acks{This research was partially supported by grant R01MH118927 of the National Institutes of Health, as well as funding from SAMSI.}

\newpage

\vskip 0.2in
\bibliography{Fiedler}

\begin{thebibliography}{49}
\providecommand{\natexlab}[1]{#1}
\providecommand{\url}[1]{\texttt{#1}}
\expandafter\ifx\csname urlstyle\endcsname\relax
  \providecommand{\doi}[1]{doi: #1}\else
  \providecommand{\doi}{doi: \begingroup \urlstyle{rm}\Url}\fi

\bibitem[Arbel et~al.(2021)Arbel, Beraha, and Bystrova]{arbel2021bayesian}
Julyan Arbel, Mario Beraha, and Daria Bystrova.
\newblock Bayesian block-diagonal graphical models via the fiedler prior.
\newblock In \emph{SFdS-52 Journ{\'e}es de Statistique de la Soci{\'e}t{\'e}
  Francaise de Statistique}, pages 1--6, 2021.

\bibitem[Bartlett and Mendelson(2002)]{bartlett2002rademacher}
Peter~L Bartlett and Shahar Mendelson.
\newblock Rademacher and gaussian complexities: Risk bounds and structural
  results.
\newblock \emph{Journal of Machine Learning Research}, 3\penalty0
  (Nov):\penalty0 463--482, 2002.

\bibitem[Batson et~al.(2012)Batson, Spielman, and Srivastava]{batson2012twice}
Joshua Batson, Daniel~A Spielman, and Nikhil Srivastava.
\newblock Twice-ramanujan sparsifiers.
\newblock \emph{SIAM Journal on Computing}, 41\penalty0 (6):\penalty0
  1704--1721, 2012.

\bibitem[Belkin and Niyogi(2003)]{belkin2003laplacian}
Mikhail Belkin and Partha Niyogi.
\newblock Laplacian eigenmaps for dimensionality reduction and data
  representation.
\newblock \emph{Neural computation}, 15\penalty0 (6):\penalty0 1373--1396,
  2003.

\bibitem[Bousquet et~al.(2003)Bousquet, Boucheron, and
  Lugosi]{bousquet2003introduction}
Olivier Bousquet, St{\'e}phane Boucheron, and G{\'a}bor Lugosi.
\newblock Introduction to statistical learning theory.
\newblock In \emph{Summer School on Machine Learning}, pages 169--207.
  Springer, 2003.

\bibitem[Boyd(2006)]{boyd2006convex}
Stephen Boyd.
\newblock Convex optimization of graph laplacian eigenvalues.
\newblock In \emph{Proceedings of the International Congress of
  Mathematicians}, volume~3, pages 1311--1319, 2006.

\bibitem[Candes et~al.(2008)Candes, Wakin, and Boyd]{candes2008enhancing}
Emmanuel~J Candes, Michael~B Wakin, and Stephen~P Boyd.
\newblock Enhancing sparsity by reweighted $\ell$1 minimization.
\newblock \emph{Journal of Fourier analysis and applications}, 14\penalty0
  (5-6):\penalty0 877--905, 2008.

\bibitem[Carmichael(2021)]{carmichael2021folded}
Iain Carmichael.
\newblock The folded concave laplacian spectral penalty learns block diagonal
  sparsity patterns with the strong oracle property.
\newblock \emph{arXiv preprint arXiv:2107.03494}, 2021.

\bibitem[Chung(1997)]{chung1997spectral}
Fan~RK Chung.
\newblock \emph{Spectral graph theory}.
\newblock Number~92. American Mathematical Soc., 1997.

\bibitem[Donath and Hoffman(1972)]{donath1972algorithms}
William~E Donath and Alan~J Hoffman.
\newblock Algorithms for partitioning of graphs and computer logic based on
  eigenvectors of connection matrices.
\newblock \emph{IBM Technical Disclosure Bulletin}, 15\penalty0 (3):\penalty0
  938--944, 1972.

\bibitem[Duchi et~al.(2011)Duchi, Hazan, and Singer]{duchi2011adaptive}
John Duchi, Elad Hazan, and Yoram Singer.
\newblock Adaptive subgradient methods for online learning and stochastic
  optimization.
\newblock \emph{Journal of machine learning research}, 12\penalty0
  (Jul):\penalty0 2121--2159, 2011.

\bibitem[Fan and Li(2001)]{fan2001variable}
Jianqing Fan and Runze Li.
\newblock Variable selection via nonconcave penalized likelihood and its oracle
  properties.
\newblock \emph{Journal of the American statistical Association}, 96\penalty0
  (456):\penalty0 1348--1360, 2001.

\bibitem[Fan et~al.(2019)Fan, Ma, and Zhong]{fan2019selective}
Jianqing Fan, Cong Ma, and Yiqiao Zhong.
\newblock A selective overview of deep learning.
\newblock \emph{arXiv preprint arXiv:1904.05526}, 2019.

\bibitem[Garey et~al.(1974)Garey, Johnson, and Stockmeyer]{garey1974some}
Michael~R Garey, David~S Johnson, and Larry Stockmeyer.
\newblock Some simplified np-complete problems.
\newblock In \emph{Proceedings of the sixth annual ACM symposium on Theory of
  computing}, pages 47--63, 1974.

\bibitem[Godsil and Royle(2013)]{godsil2013algebraic}
Chris Godsil and Gordon~F Royle.
\newblock \emph{Algebraic graph theory}, volume 207.
\newblock Springer Science \& Business Media, 2013.

\bibitem[Golowich et~al.(2018)Golowich, Rakhlin, and Shamir]{golowich2018size}
Noah Golowich, Alexander Rakhlin, and Ohad Shamir.
\newblock Size-independent sample complexity of neural networks.
\newblock In \emph{Conference On Learning Theory}, pages 297--299. PMLR, 2018.

\bibitem[Graves(2013)]{graves2013generating}
Alex Graves.
\newblock Generating sequences with recurrent neural networks.
\newblock \emph{arXiv preprint arXiv:1308.0850}, 2013.

\bibitem[Hagen and Kahng(1992)]{hagen1992new}
Lars Hagen and Andrew~B Kahng.
\newblock New spectral methods for ratio cut partitioning and clustering.
\newblock \emph{IEEE transactions on computer-aided design of integrated
  circuits and systems}, 11\penalty0 (9):\penalty0 1074--1085, 1992.

\bibitem[Hall(1970)]{hall1970r}
Kenneth~M Hall.
\newblock An r-dimensional quadratic placement algorithm.
\newblock \emph{Management science}, 17\penalty0 (3):\penalty0 219--229, 1970.

\bibitem[Hinton et~al.(2012)Hinton, Srivastava, Krizhevsky, Sutskever, and
  Salakhutdinov]{hinton2012improving}
Geoffrey~E Hinton, Nitish Srivastava, Alex Krizhevsky, Ilya Sutskever, and
  Ruslan~R Salakhutdinov.
\newblock Improving neural networks by preventing co-adaptation of feature
  detectors.
\newblock \emph{arXiv preprint arXiv:1207.0580}, 2012.

\bibitem[Horn and Johnson(2012)]{horn2012matrix}
Roger~A Horn and Charles~R Johnson.
\newblock \emph{Matrix analysis}.
\newblock Cambridge university press, 2012.

\bibitem[Horn et~al.(1998)Horn, Rhee, and Wasin]{horn1998eigenvalue}
Roger~A Horn, Noah~H Rhee, and So~Wasin.
\newblock Eigenvalue inequalities and equalities.
\newblock \emph{Linear Algebra and its Applications}, 270\penalty0
  (1-3):\penalty0 29--44, 1998.

\bibitem[Jiang and Lin(2018)]{jiang2018graph}
Bo~Jiang and Doudou Lin.
\newblock Graph laplacian regularized graph convolutional networks for
  semi-supervised learning.
\newblock \emph{arXiv preprint arXiv:1809.09839}, 2018.

\bibitem[Kingma and Ba(2014)]{kingma2014adam}
Diederik~P Kingma and Jimmy Ba.
\newblock Adam: A method for stochastic optimization.
\newblock \emph{arXiv preprint arXiv:1412.6980}, 2014.

\bibitem[Kipf and Welling(2016)]{kipf2016semi}
Thomas~N Kipf and Max Welling.
\newblock Semi-supervised classification with graph convolutional networks.
\newblock \emph{arXiv preprint arXiv:1609.02907}, 2016.

\bibitem[Krogh and Hertz(1992)]{krogh1992simple}
Anders Krogh and John~A Hertz.
\newblock A simple weight decay can improve generalization.
\newblock In \emph{Advances in neural information processing systems}, pages
  950--957, 1992.

\bibitem[Langford et~al.(2009)Langford, Li, and Zhang]{langford2009sparse}
John Langford, Lihong Li, and Tong Zhang.
\newblock Sparse online learning via truncated gradient.
\newblock In \emph{Advances in neural information processing systems}, pages
  905--912, 2009.

\bibitem[Ledoux and Talagrand(2013)]{ledoux2013probability}
Michel Ledoux and Michel Talagrand.
\newblock \emph{Probability in Banach Spaces: isoperimetry and processes}.
\newblock Springer Science \& Business Media, 2013.

\bibitem[Massart(2000)]{massart2000some}
Pascal Massart.
\newblock Some applications of concentration inequalities to statistics.
\newblock In \emph{Annales de la Facult{\'e} des sciences de Toulouse:
  Math{\'e}matiques}, volume~9, pages 245--303, 2000.

\bibitem[Neyshabur et~al.(2017)Neyshabur, Bhojanapalli, and
  Srebro]{neyshabur2017pac}
Behnam Neyshabur, Srinadh Bhojanapalli, and Nathan Srebro.
\newblock A pac-bayesian approach to spectrally-normalized margin bounds for
  neural networks.
\newblock \emph{arXiv preprint arXiv:1707.09564}, 2017.

\bibitem[Ng et~al.(2001)Ng, Jordan, and Weiss]{ng2001spectral}
Andrew Ng, Michael Jordan, and Yair Weiss.
\newblock On spectral clustering: Analysis and an algorithm.
\newblock \emph{Advances in neural information processing systems}, 14, 2001.

\bibitem[Petersen et~al.(2008)Petersen, Pedersen, et~al.]{petersen2008matrix}
Kaare~Brandt Petersen, Michael~Syskind Pedersen, et~al.
\newblock The matrix cookbook.
\newblock \emph{Technical University of Denmark}, 7\penalty0 (15):\penalty0
  510, 2008.

\bibitem[Polson and Ro{\v{c}}kov{\'a}(2018)]{polson2018posterior}
Nicholas~G Polson and Veronika Ro{\v{c}}kov{\'a}.
\newblock Posterior concentration for sparse deep learning.
\newblock In \emph{Advances in Neural Information Processing Systems}, pages
  930--941, 2018.

\bibitem[Ruder(2016)]{ruder2016overview}
Sebastian Ruder.
\newblock An overview of gradient descent optimization algorithms.
\newblock \emph{arXiv preprint arXiv:1609.04747}, 2016.

\bibitem[Scardapane et~al.(2017)Scardapane, Comminiello, Hussain, and
  Uncini]{scardapane2017group}
Simone Scardapane, Danilo Comminiello, Amir Hussain, and Aurelio Uncini.
\newblock Group sparse regularization for deep neural networks.
\newblock \emph{Neurocomputing}, 241:\penalty0 81--89, 2017.

\bibitem[Shalev-Shwartz and Ben-David(2014)]{shalev2014understanding}
Shai Shalev-Shwartz and Shai Ben-David.
\newblock \emph{Understanding machine learning: From theory to algorithms}.
\newblock Cambridge university press, 2014.

\bibitem[Spielman(2019)]{Spielman2019Algebraic}
Daniel Spielman.
\newblock Spectral and algebraic graph theory.
\newblock 2019.

\bibitem[Srivastava et~al.(2014)Srivastava, Hinton, Krizhevsky, Sutskever, and
  Salakhutdinov]{srivastava2014dropout}
Nitish Srivastava, Geoffrey Hinton, Alex Krizhevsky, Ilya Sutskever, and Ruslan
  Salakhutdinov.
\newblock Dropout: a simple way to prevent neural networks from overfitting.
\newblock \emph{The journal of machine learning research}, 15\penalty0
  (1):\penalty0 1929--1958, 2014.

\bibitem[Sun et~al.(2006)Sun, Boyd, Xiao, and Diaconis]{sun2006fastest}
Jun Sun, Stephen Boyd, Lin Xiao, and Persi Diaconis.
\newblock The fastest mixing markov process on a graph and a connection to a
  maximum variance unfolding problem.
\newblock \emph{SIAM review}, 48\penalty0 (4):\penalty0 681--699, 2006.

\bibitem[Tam and Dunson(2020)]{tam2020fiedler}
Edric Tam and David Dunson.
\newblock Fiedler regularization: Learning neural networks with graph sparsity.
\newblock \emph{arXiv preprint arXiv:2003.00992}, 2020.

\bibitem[Tibshirani(1996)]{tibshirani1996regression}
Robert Tibshirani.
\newblock Regression shrinkage and selection via the lasso.
\newblock \emph{Journal of the Royal Statistical Society: Series B
  (Methodological)}, 58\penalty0 (1):\penalty0 267--288, 1996.

\bibitem[Tuck et~al.(2019)Tuck, Hallac, and Boyd]{tuck2019distributed}
Jonathan Tuck, David Hallac, and Stephen Boyd.
\newblock Distributed majorization-minimization for laplacian regularized
  problems.
\newblock \emph{IEEE/CAA Journal of Automatica Sinica}, 6\penalty0
  (1):\penalty0 45--52, 2019.

\bibitem[Vladimirova et~al.(2018)Vladimirova, Verbeek, Mesejo, and
  Arbel]{vladimirova2018understanding}
Mariia Vladimirova, Jakob Verbeek, Pablo Mesejo, and Julyan Arbel.
\newblock Understanding priors in bayesian neural networks at the unit level.
\newblock \emph{arXiv preprint arXiv:1810.05193}, 2018.

\bibitem[Von~Luxburg(2007)]{von2007tutorial}
Ulrike Von~Luxburg.
\newblock A tutorial on spectral clustering.
\newblock \emph{Statistics and computing}, 17:\penalty0 395--416, 2007.

\bibitem[Wan et~al.(2013)Wan, Zeiler, Zhang, Le~Cun, and
  Fergus]{wan2013regularization}
Li~Wan, Matthew Zeiler, Sixin Zhang, Yann Le~Cun, and Rob Fergus.
\newblock Regularization of neural networks using dropconnect.
\newblock In \emph{International conference on machine learning}, pages
  1058--1066, 2013.

\bibitem[Weinstein et~al.(2013)Weinstein, Collisson, Mills, Shaw, Ozenberger,
  Ellrott, Shmulevich, Sander, Stuart, Network, et~al.]{weinstein2013cancer}
John~N Weinstein, Eric~A Collisson, Gordon~B Mills, Kenna R~Mills Shaw, Brad~A
  Ozenberger, Kyle Ellrott, Ilya Shmulevich, Chris Sander, Joshua~M Stuart,
  Cancer Genome Atlas~Research Network, et~al.
\newblock The cancer genome atlas pan-cancer analysis project.
\newblock \emph{Nature genetics}, 45\penalty0 (10):\penalty0 1113, 2013.

\bibitem[Zeng et~al.(2019)Zeng, Pang, Sun, and Cheung]{zeng2019deep}
Jin Zeng, Jiahao Pang, Wenxiu Sun, and Gene Cheung.
\newblock Deep graph laplacian regularization for robust denoising of real
  images.
\newblock In \emph{Proceedings of the IEEE Conference on Computer Vision and
  Pattern Recognition Workshops}, pages 0--0, 2019.

\bibitem[Zhang(2010)]{zhang2010nearly}
Cun-Hui Zhang.
\newblock Nearly unbiased variable selection under minimax concave penalty.
\newblock 2010.

\bibitem[Zou(2006)]{zou2006adaptive}
Hui Zou.
\newblock The adaptive lasso and its oracle properties.
\newblock \emph{Journal of the American statistical association}, 101\penalty0
  (476):\penalty0 1418--1429, 2006.

\end{thebibliography}

\end{document}